\documentclass[11pt,a4paper]{article}
\usepackage[hyperref]{AACL-IJCNLP2020}
\usepackage{times}
\usepackage{latexsym}

\usepackage{microtype}
\usepackage{multirow}
\usepackage{url}
\usepackage{amsmath}
\usepackage{algorithm}
\usepackage{algorithmic}
\usepackage{graphicx}

\aclfinalcopy 

\title{Towards Controlled and Diverse Generation of Article Comments}

\author{Linhao Zhang\textsuperscript{\rm 1},  Houfeng Wang\textsuperscript{\rm 1}\\ 
\textsuperscript{\rm 1}MOE Key Lab of Computational Linguistics, Peking University, Beijing, 100871, China\\ 
\textsuperscript{\rm 2}Baidu Inc., China\\
\{zhanglinhao, wanghf\}@pku.edu.cn\\
}

\date{}

\begin{document}
\maketitle
\begin{abstract}

Much research in recent years has focused on automatic article 
commenting. However, few of previous studies focus on the controllable generation 
of comments. Besides, they tend to 
generate dull and commonplace comments, which further limits their practical 
application. In this paper, we make the first step towards controllable generation of comments, by building a system that can explicitly control the emotion of the generated comments. To achieve this, we associate each kind of emotion category with an embedding and adopt a dynamic fusion mechanism to fuse this embedding 
into the decoder. A sentence-level emotion classifier is further employed 
to better guide the model to generate comments expressing the desired emotion. To increase the diversity of the generated 
comments, we propose a hierarchical copy mechanism that allows our model to directly 
copy words from the input articles.
We also propose a restricted beam search (RBS)
algorithm to increase intra-sentence diversity.
Experimental results show that our model can generate informative and diverse
comments that express the desired emotions with high accuracy.
\end{abstract}

\section{Introduction}
Automatic article commenting is a valuable yet challenging task. 
It requires the machine first to understand the articles and then generate coherent comments. The task was formally proposed by \cite{qin2018automatic}, along with a large-scale dataset. Much research has since focused on this task \cite{lin2018learning,Ma2018UnsupervisedMC,li2019coherent}. 

The ability to generate comments is especially useful for online news platforms, 
for the comments can encourage user engagement and interactions 
\cite{qin2018automatic}. Besides, an automatic commenting system also enables us to 
build a comment-assistant which can generate some candidate comments for users, who may later select one and refine it \cite{zheng2018automatic}.

In addition to the practical importance of this task, it also has
significant research value. It can be seen as a natural language generation 
(NLG) task, yet unlike machine translation or text summarization, the 
comments can be rather diverse. That is, for the same article, 
there can be numerous appropriate comments that are from  different angles. 
In this sense, this task is similar to dialog system, yet because the input article 
is much longer and more complex than dialog text, it is hence more challenging. 

\begin{figure}[t]
\centering
\hspace*{-0.5cm}  
\includegraphics[width=7.5cm]{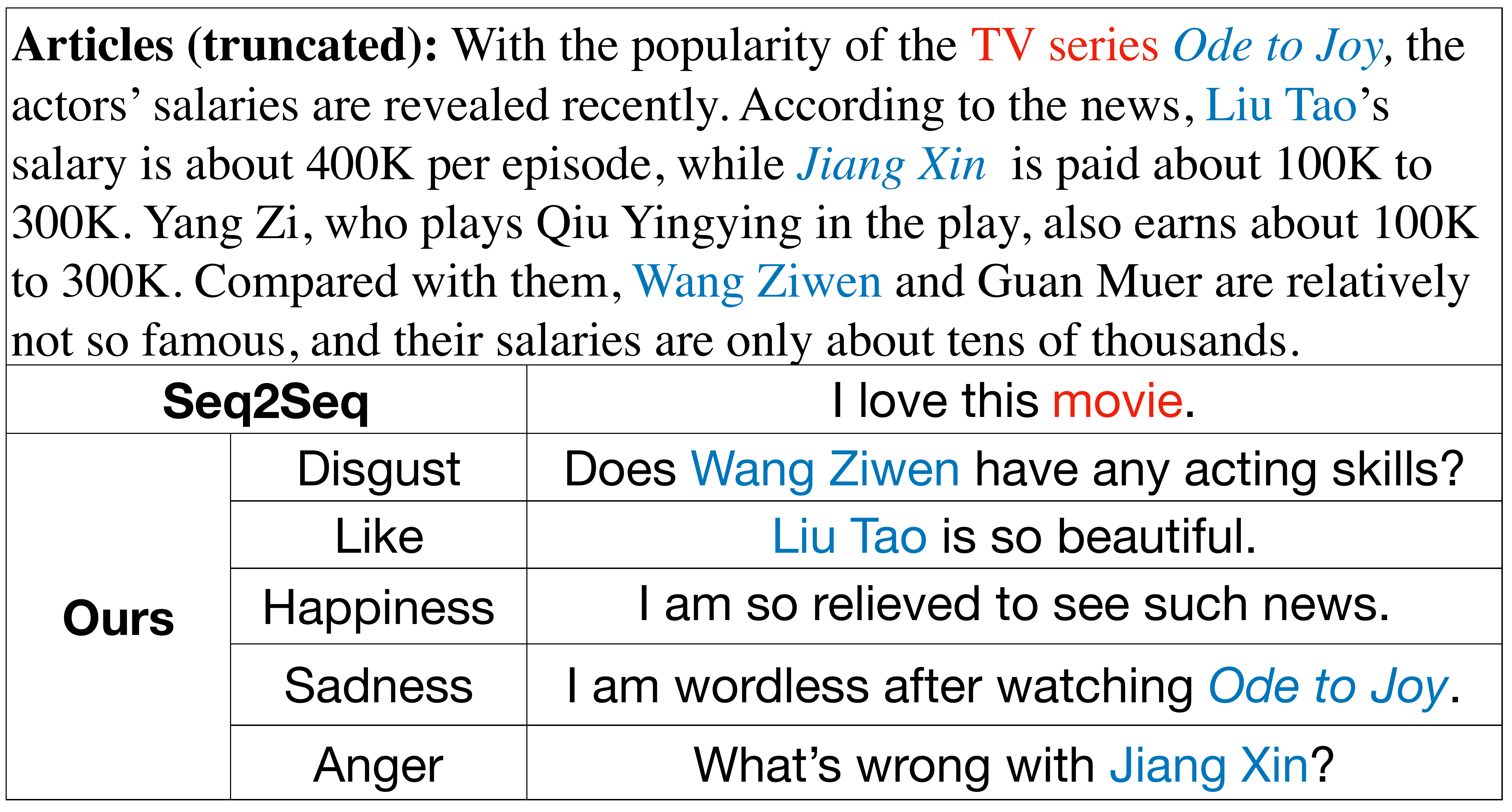}
\caption{Comparison of output comments of the Seq2Seq baseline and our model. We can 
see that the baseline cannot control the expressed emotion, and the generated comment 
is dull and irrelevant to the article (red colored). By contrast, Our model is 
emotion-controllable and can generate more diverse and relevant comments, thanks to 
the hierarchical copy mechanism (blue colored).}
\centering
\label{figure-example}
\end{figure}

Despite the importance of this task, it is still relatively new and not well-studied. 
One of the major limitations of current article commenting systems 
is that the generation process is not controllable, meaning that the 
comments are conditioned entirely on the articles, and users can not 
further control the features of comments. In this work, we make the 
first step towards controllable article commenting and propose a model to control the emotion of generated 
comments, which has wide practical application.
Take comment-assistant \cite{zheng2018automatic} as an example, it 
is far more desirable to have candidate comments each expresses a different emotion, and users can hence choose one that matches their 
own emotion towards the article. 

Another problem of current commenting systems arises from the 
limitation of the Seq2Seq framework \cite{Sutskever2014}, which has been known to suffer
from generating dull and responses that are irrelevant to the input articles \cite{li2015diversity,wei2019neural,shao2017generating}. 
As shown in Figure \ref{figure-example}, the Seq2Seq baseline 
generates \emph{I love this movie} for the input article, despite the fact that \emph{
Ode of joy} is not a movie, but a TV series.

In this work,  we propose a controllable article commenting system that can generate diverse and relevant comments. 
We first create 
two emotional datasets based on 
\cite{qin2018automatic}.  The first one is a find-grained dataset 
that contains \{Anger, Disgust, Like, Happiness and Sadness \} and 
the second one is a coarse-grained one that contains only \{Positive and Negative\}.
To incorporate the emotion information into the decoding process, we 
first associate each kind of emotion with an embedding and then feed 
the emotion embedding into the decoder at 
each time step. A dynamic fusion mechanism is employed to utilize 
the emotion information selectively at each decoding 
step. Besides, the decoding process is further guided by a  
sequence-level emotion loss term to increase the intensity of 
emotional expression.


To generate diverse comments, we propose a hierarchical 
copy mechanism to copy words from the input articles.
This is based on the observation that we can discourage the 
generation of dull comments like \emph{I don't know} by enhancing 
the relevance between comments and articles. In this way, the inter-sentence diversity 
gets increased. Moreover, we observe that the repetition problem of Seq2Seq framework 
\cite{see2017get,lin2018global} can be seen as a lack of intra-sentence diversity, 
and we further adopt a restricted beam search (RBS) algorithm to tackle this problem.


\begin{figure*}[t!]
\centering
\includegraphics[scale=0.35]{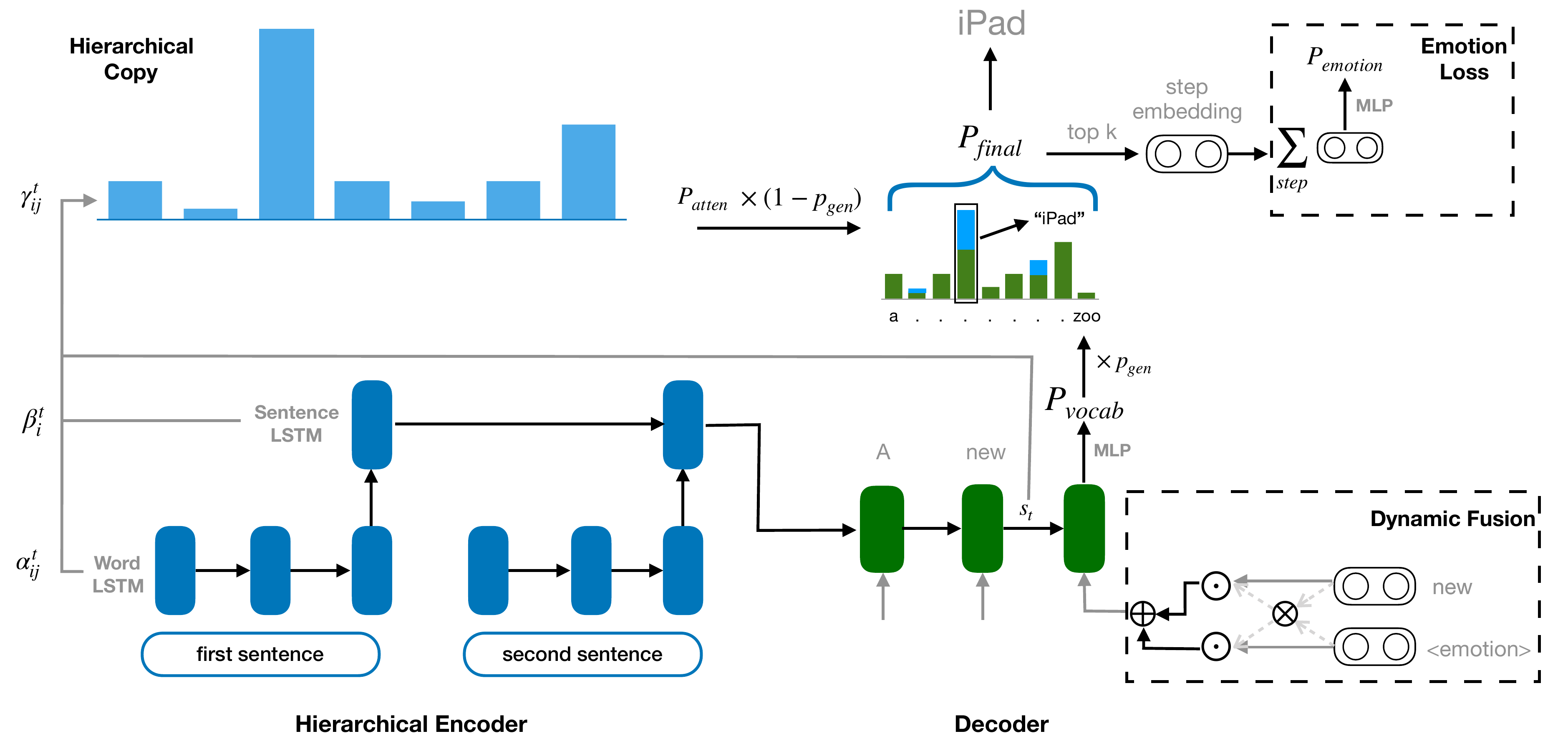}
\caption{The architecture of CCS. On the left 
is the encoder, which 
encodes articles at the word- and sentence- level. On the right 
is the decoder, with emotion information dynamically fused into
the decoding process. We add a emotion loss item to further bias the decoding 
process. Besides, a hierarchical copy mechanism is proposed to improve the diversity of the generated comments.}
\label{figure-model}
\end{figure*}
To sum up, our contributions are two-folds:

1) We make the first step to build a controllable article commenting 
system by injecting emotion into the decoding process. Features other than emotion can be controlled in a similar way.

2) We propose the hierarchical copy mechanism  and RBS algorithm to increase the inter- and intra-sentence diversity, respectively.
\section{Method}

\label{section:model}

The overall architecture of our proposed model, Controllable Commenting System (CCS), is shown in 
Figure \ref{figure-model}. We describe the details in the following 
subsections.

\subsection{Task Defination}
The task can be formulated as follows: Given an article $D$ and an 
emotion category $\chi$, the goal is to generate an appropriate comment $Y$ that expresses the emotion $\chi$.

Specifically, $D$ consists of $N_D$ sentences. Each 
sentence consists of $T$ words, where $T$ can vary among different 
sentences. For each word, we use $e_i$ to denote its word embedding.

\subsection{Basic Structure: Hierarchical Seq2Seq}
\label{encoder}
We encode an article by first 
building representations of sentences and then aggregating them into an article representation. 

\textbf{Word-Level Encoder} -
We first encode the article $D$ on the word level. For each sentence in D, the word-level LSTM encoder ${LSTM}_w^{enc}$ reads the words sequentially and updates its hidden state $h_i$. Formally,
\begin{equation}
h_i = {LSTM}_{w}^{enc}(e_i,h_{i-1})
\end{equation}
The last hidden state, $h_T$, stores information of the whole sentence and thus
can be used to represent that sentence:
\begin{equation}
\hat{e} = h_T
\end{equation}

\textbf{Sentence-Level Encoder} -
Given sentence embeddings $(\hat{e}_1,\hat{e}_2,...,\hat{e}_{N_D})$, we then encode 
the article at the 
sentence-level:
\begin{equation}
g_i = LSTM_s^{enc}(\hat{e}_i,g_{i-1})
\label{equation:sent_enc}
\end{equation}
where the $LSTM_s^{enc}$ is the sentence-level LSTM encoder and $g_i$ is its hidden state.
The last hidden state, $g_{N_D}$, aggregates information of all the sentences and is later used by the decoder to initialize its hidden state.

\label{subsection:encoder}

\textbf{Decoder}
\label{sect-decoder}
Similarly, the decoder LSTM ${LSTM}^{dec}$ updates its hidden state $s_t$ at time step t:

\begin{equation}
s_t = LSTM^{dec}(\overline{e}_t,s_{t-1})
\label{equation-decoder}
\end{equation}

where $\overline{e}_t$ is the embedding of the previous word. While 
training, this is the previous word of 
the reference comment; at test time it is the previous word emitted by the decoder.
The attention mechanism \cite{luong2015effective} is also applied.
 Formally,
At each decoding step, we compute the attention weights between the 
current decoder hidden vector and all the outputs of the 
sentence-level encoder. We then compute the context vector $c_{t}$ 
using the weighted sum of the sentence-level encoder outputs and 
obtain our attention vector $a_t$ for the final prediction. Formally,
\begin{equation}
\begin{aligned}
&m_{t,i} = softmax(s_t^TW_gg_i) \\
&c_{t} = \sum_{i}m_{t,i}g_i \\
&a_t = tanh(W_a[c_t;s_t] + b_a)\\
& P_{\mathrm{vocab}}=\operatorname{softmax}(W_pa_t)
\end{aligned}
\end{equation}
where $W_a$ ,$W_g$, $W_p$ and $b_a$ are model parameters; $[;]$ is
vector concatenation.

\subsection{Dynamic Fusion Mechanism}
To control the emotions expressed by the comments, we first associate each emotion with 
an embedding $v_\chi$. For clarification, we note that $\chi$ is the 
user-specified emotion category and $v_\chi$ is obtained by a trainable embedding 
layer. We then add the emotion vector $v_\chi$ to the decoder input $\overline{e}_t$ at each decoding step, changing Equation \ref{equation-decoder} into:

\begin{equation}
s_t = LSTM^{dec}(v_\chi + \overline{e}_t,s_{t-1})
\label{euqation-simple}
\end{equation}


In this way, emotion information is injected into the decoding 
process at each decoding step.
However, this method is overly simplistic. It uses emotion information indiscriminately at 
each decoding step, yet not all steps require the same 
amount of emotion information. 
Simply using the same emotion embedding during the whole generation process may sacrifice grammaticality \cite{ghosh2017affect}. 

To solve this issue, we adopt a dynamic fusion mechanism that 
can update and selectively utilize the emotion embedding at 
each decoding step. Besides, given that the same word can convey different emotions, we also alter the word embedding according to emotions to be expressed. Specifically, we first employ a emotion gate $z_t^e$ and a word 
gate $z_t^w$: 
\begin{equation}
z_{t}^e=\sigma\left(W_e[v_\chi;s_{t-1}]+b_e\right)
\label{equation-readgate}
\end{equation}

\begin{equation}
z_{t}^w=\sigma\left(W_w[v_\chi;s_{t-1}]+b_w\right)
\label{equation-readgate}
\end{equation}

where $\sigma$ is the sigmoid function; $W_e$, $W_w$, $b_e$, $b_w$ are model parameters. The resulting vectors $z_t^e$ and $z_t^w$ are of the same 
dimension as $v_\chi$ and $s_{t-1}$. They are used to control the 
amount of emotion information used in current decoding step, changing
Equation \ref{euqation-simple} into:

\begin{equation}
s_t = LSTM^{dec}(v_\chi \odot z_t^e + \overline{e}_t \odot z_t^w,s_{t-1})
\end{equation}
where $\odot$ is element-wise multiplication.





Experiments show that the dynamic fusion mechanism leads to 
better generation quality and higher emotion accuracy.

\subsection{Emotion Loss}
To further guide the model to generate comments expressing the 
desired emotion, we use a sequence-level emotion classifier \cite{song2019generating} to guide the generation process.

We first approximate the step embedding $E_t$ using the weighted sum 
of embeddings of words with the \emph{TopK} probabilities, where $K$ 
is a hyperparameter.

\begin{equation}
E_t =\sum_{w_{i} \in Top K} P\left(w_{i}\right) \cdot E m b\left(w_{i}\right)
\end{equation}

Then, we calculate the emotion distribution using the 
average step embeddings of all time steps, based on which we 
calculate the final emotion classification loss:

\begin{equation}
G(E | Y)=\operatorname{softmax}(W_g \cdot \frac{1}{T} \sum_{t=1}^{T} E_t))
\end{equation}
\begin{equation}
\mathcal{L}_{emo}=-P(E) \log (G(E | Y))
\label{equation-cla}
\end{equation}
where $W_g$ is model parameters and P(E) is the gold emotion distribution.

Intuitively, the emotion loss forces a strong dependency relationship between the input emotion category and generated comment, further biasing the comments towards our desired emotion.

\subsection{Hierarchical Copy Mechanism}
A key limitation of Seq2Seq models is that they tend to generate dull and commonplace 
text such as \emph{I don't know} and \emph{I like it}. We observe that we can 
encourage the generation of diverse comments by enhancing the relevance between 
comments and source articles. 

To achieve this, we adopt copy mechanism \cite{see2017get}, which allows for copying words from source text. We adapt the original copy mechanism to consider the hierarchical structure
of articles. Specifically, we first compute attention $\beta_{t}$ over sentences.

\begin{equation}
\beta_{i}^{t}=\operatorname{softmax}\left(s_{t}^TW_xg_{i}\right)
\end{equation}
where $W_x$ is model parameter, $s_{t}$ is decoder's $t_{th}$ hidden state and $g_{i}$ is $i_{th}$ sentence representation.

We then compute the attentions over individual words and normalize them using the sentence-level attention $\beta_{t}$.

\begin{equation}
\alpha_{i j}^{t}=softmax(s_{t} W_{y} h_{i}^{j})
\end{equation}
\begin{equation}
\gamma_{i j}^{t}=\beta_{i}^{t} * \alpha_{ij}^t 
\end{equation}
where $W_y$ is model parameters, $h^j_i$ is the $j_{th}$ words of 
the $i_{th}$ sentence. In this way, words from more informative sentences are rewarded, and words from less informative ones are discouraged.

Finally, we get the context vector $c_t$ 
using the weighted sum of all the words and obtain our 
attention vector at for the final prediction. Formally,
\begin{equation}
\begin{aligned} &c_{t}=\sum_{i, j} \gamma_{i j}^{t} h_{i}^{j} \\ &a_{t} =\tanh \left(W_{a}\left[c_{t} ; s_{t}\right]+b_{a}\right) \\&P_{\text {vocab }} =\operatorname{softmax}\left(a_{t}\right) \end{aligned}
\end{equation}
where \(W_{a}, b_{a}\) are model parameters. \(P_{v o c a b}\) is a probability
distribution over all words in the vocabulary. 
Then, the attention distribution $\gamma_{ij}^t$ and the vocabulary distribution $P_{vocab}$ are then weighted and summed to obtain the final word distribution:
\begin{equation}
p_{\mathrm{gen}}=\sigma\left(w_{a}^{T} a_{t}+w_{s}^{T} s_{t}+w_{e}^{T} \overline{e}_{t}+b_{\mathrm{ptr}}\right)
\end{equation}
\begin{equation}
P(w)=p_{\mathrm{gen}} P_{\mathrm{vocab}}(w)+\left(1-p_{\mathrm{gen}}\right) \sum_{i j : w_{i j}=w} \gamma_{i j}^{t}
\end{equation}

Intuitively, $p_{gen}$ can be seen as a soft switch to choose between generating a 
word from the vocabulary, and copying a word from the source article. Experiment 
results show that our hierarchical copy mechanism can effectively improve comments 
diversity.

\subsection{Restricted Beam Search}
The above method can improve the inter-sentence diversity
of comments. However, the Seq2Seq framework has also long been known of generating repeating and redundant words and phrases, as the example shown in \cite{lin2018global}:

\emph{Fatah \textbf{officially officially} elects Abbas as \textbf{candidate} for \textbf{candidate}}.

We regard this problem as a lack of intra-sentence diversity.
To mitigate this issue, we propose the Restricted Beam Search (RBS) algorithm. Based on the observation that ground truth comments 
seldom contain the same n-grams multiple times, we explicitly lower 
the probability of those words that would create repetitive n-grams 
at each decoding step. 
Specifically, we will lower the probability of generating $w_t$
 by $\eta$ if outputting $w_t$ creates
an n-gram that already exists in the previously decoded sequence of 
the current beam. Formally, the probability of $w_t$ will be modified as the following.

\begin{small}
\begin{equation}
P(w_t)=\left\{\begin{array}{ll}{P(w_t)} & {\text {if } w_t \text{ won't create}} \\ &{\text{ 	\text{ \text{	repetitive n-gram}}}}\\ 
{max(P(w_t) - \pi\eta,0)} & {\text{otherwise}}\end{array}\right.
\end{equation}
\end{small}
where $\eta$ is a hyperparameter and $\pi$ is the times that $w_t$-created n-gram has occurred in the previously decoded sequence of the current beam.

Despite its brevity, we found that this procedure can substantially improve intra-sentence diversity.  Note that the RBS algorithm is notably different from \cite{paulus2017deep}. They 
set $P\left(w_t\right) = 0$ during beam search, when outputting $w_t$ would create a n-gram 
that already exists. We, on the other hand, lower the probability of $w_t$ by a certain amount. We believe that forcing the decoder to 
\emph{never} output the same n-gram more than once is too strict, for repetitive trigrams do exist in natural language.
In this sense, our RBS algorithm can be seen as a soft version of \cite{paulus2017deep} that
enables more flexibility.

\section{Experiments}

\label{sect-experiments}
\subsection{Dataset}
Currently there is no off-the-shelf article commenting dataset annotated with emotions, so we created our own datasets based on the Tencent News dataset released by \cite{qin2018automatic}. For fine-grained setting, which involves
\emph{\{Anger, Disgust, Like, Happiness and Sadness\}}, we trained a Bi-LSTM emotion tagger using
NLPCC$\footnote{http://tcci.ccf.org.cn/conference/2014/}$ dataset. 
For coarse-grained setting, which involves \emph{\{Positive and Negative\}}, we used
a well-trained emotion tagger provided by 
Baidu$\footnote{https://ai.baidu.com/tech/nlp/sentiment\_classify}$, 
which provides the function to classify a text into \emph{positive} 
or \emph{negative} with high accuracy.
We then annotated the Tencent News dataset with the two taggers, 
creating two emotional commenting datasets, called Tencent-coarse and 
Tencent-fine respectively$\footnote{We will release the two datasets to promote future research}$.

For the original Tencent News dataset, there are 169,060 articles in the training set, 4,455 in the development set and 1,397 in the test set. There are several comments for the same article, and many of these comments are short and meaningless. To promote the quality of the generated comments, we trained our model with only one comment that has the most upvotes for each article. We found this cleaning procedure led to better generation quality and expedited training substantially.

\begin{table*}[t!]
\begin{center}
\scalebox{0.9}{
\begin{tabular}{l|ccc|ccc|cc}
\hline
\multicolumn{1}{c|}{\multirow{2}*{\textbf{Model}}} 
& \multicolumn{3}{|c|}{\textbf{Generation Quality}} 
& \multicolumn{3}{|c|}{\textbf{Diversity}} 
& \multicolumn{2}{|c}{\textbf{Emotion Acc.}} \\ 
\cline{2-9}
                     & BLEU &METEOR & ROUGE-L&D1&D2&D3  & Acc-C & Acc-F\\ \hline 
Seq2Seq \cite{Sutskever2014}             & 5.15        & 9.43        & 22.67  &4.51&16.48&23.56 &-&-    \\  
Transformer  \cite{vaswani2017attention}      & 4.23       & 9.89   & 24.09  &3.98&15.23&21.96& -   &- \\   
Flat-Emo  \cite{Huang2018} & 4.13 & 7.75 &23.40 &3.38&10.85&15.99&76.21&43.92 \\
Hier-Emo \cite{Huang2018}  &5.02&8.97&24.14&3.54&13.01&20.25&78.35&44.26\\ \hline \hline
CCS                    & \textbf{7.54}  & \textbf{10.50} & \textbf{25.29} &\textbf{5.36}&\textbf{19.74}&\textbf{29.36}& \textbf{83.10} &\textbf{48.34} \\ \hline

\end{tabular}
}
\end{center}
\caption{\label{table-main} Results of the generation quality,
diversity and emotion accuracy(\%).} 
\label{table-bleu}
\end{table*}

\label{sect-baidu}
\subsection{Training Procedure}
We trained our models on a single NVIDIA TITAN RTX GPU. 
The LSTMs \cite{hochreiter1997long} are with 512-dimensional hidden states.  The dimensions of both word embeddings and sentiment embeddings are 512.
Dropout \cite{hinton2012improving} is used with dropout rate set to 0.3. The number of 
layers of LSTM encoder/decoder is set to 2. The batch size is set to 64.

We use Adam 
\cite{kingma2014adam} with
learning rate = $10^{-3}$, $ \beta_{1}=0.9, \beta_{2}=0.98$ and $\epsilon=10^{-9}$. 
 We obtain the pretrained word embedding by 
training an unsupervised word2vec \cite{mikolov2013efficient} model on the training set. 

We compare the char-based model with the word-based model. For the latter we use jieba$\footnote{https://github.com/fxsjy/jieba}$ for Chinese word segmentation (CWS) to preprocess the text. We find that char-based model given better performance, so we adopt this setting. This results is in line with \cite{Meng2019IsWS}.

We shared the vocabulary between articles and comments, and the 
vocabulary size was restricted to 5,000. During training, we truncated
the article to 30 sentences and limited the length of each sentence to 
80 tokens. At test time our comments were produced using restricted beam search with beam size = 5. We found that setting n to 1 and $\eta$ to 0.5 is enough for effective repetition reduction.  $\xi$, $\tau$, N, K are set to  0.01, 1, 20 and 50 respectively.

\subsection{Systems for Comparison}
As we have mentioned, this is the first work to consider the emotion 
factor for article commenting. We did not find closely-related 
baselines in the literature. Nevertheless, we first chose two baselines that are widely used in NLG tasks:
\begin{itemize}

\item  \textbf{Seq2Seq -}  In this paper we use Seq2Seq \cite{Sutskever2014} model enhanced with attention mechanism \cite{luong2015effective}.

\item  \textbf{Transformer -}  The Transformer \cite{vaswani2017attention} also follows the encoder-decoder paradigm but relies on self-attention instead of RNN. 
\end{itemize}

However, these two baselines cannot control the expressed emotion. 
\cite{Huang2018} proposed to generate dialogue with expressed emotion, we adapted their approach and created
two new models that are emotion-controllable:

\begin{itemize}
\item  \textbf{Flat-Emo} -  The article is represented as a flat structure, and the emotion embedding serves as an input to every decoding step to a Seq2Seq network, as in \cite{Huang2018}.

\item  \textbf{Hier-Emo} - Similar to \textbf{Flat-Emo}, only that the article is encoded in a hierarchical manner.

\end{itemize}

\section{Results}

\subsection{Generation Quality}
Automatic metrics such as BLEU, METEOR, and ROUGE are widely used for NLG tasks \cite{lin2018global,Rose2016,song2019generating}. We adopt these metrics to evaluate generation quality, that is, whether the comments are relevant and grammatical.

The results can be seen in Table 
\ref{table-main}. The first
observation is that the first three baselines give similar and 
unsatisfactory results. This may due to the general limitations of 
non-hierarchical structure. Hier-Emo, on the other hand, makes use 
of this structure information and hence gives better performance. 
However, it still suffers from the general problems of Seq2Seq 
framework. Besides, the 
emotion information is indiscriminately fused into the decoding 
process, which also hurts generation quality \cite{ghosh2017affect}.

Compared to baselines, our model gives substantially better performance in terms of all metrics. One reason is that we adopt the hierarchical copy mechanism, which improves the coherence between comments and articles.
Besides, by adopting the dynamic fusion mechanism, our model can  use the emotion information selectively, which is also beneficial to 
generation quality.

For clarification, we note that at test time the input emotion tags 
are not used as external knowledge. That is, 
the specified emotion categories are manually designed rather 
than reflecting the emotions of the gold comments, so
the comparison with models that cannot use emotion information (e.g., Seq2Seq and Transformer) would be fair. 
The reported results are
averaged over all emotion categories.
Although this setting makes the task harder, we believe it is much closer to practical 
scenarios.

\begin{table*}[t!]
\begin{center}
\begin{tabular}{l|ccc|ccccccc}
\hline
\multicolumn{1}{c|}{\multirow{2}*{\textbf{Model}}} 
& \multicolumn{3}{|c|}{\textbf{Coarse-grained}} 
& \multicolumn{6}{|c}{\textbf{Fine-grained}} 
\\ \cline{2-10}
& Pos. &Neg. & Ave.&Disgust&Anger&Sadness& Like& Happiness & Ave.\\ \hline 
CCS-Emo  & 79.89 &77.09 &78.04 &59.24&31.38&41.02&63.54&39.58&46.95\\ 
CCS  & 81.66 &84.54 &\textbf{83.10} & 61.39&33.52&43.19&69.72&33.88&\textbf{48.34} \\ \hline

\end{tabular}

\end{center}
\caption{\label{table-emotion} Effectiveness of our emotion-control approach (\%).} 
\end{table*}

\begin{table}[t!]
\begin{center}
\begin{tabular}{l|ccc}
\hline
\textbf{Model}        &\textbf{D1}&\textbf{D2} &\textbf{D3}   \\ \hline 
CCS w/o RBS & 3.28 & 14.25 & 22.32   \\
CCS w/o HC & 4.05 & 17.80 & 29.17   \\
CCS       &\textbf{5.36}&\textbf{19.74}&\textbf{29.36} \\ \hline
\end{tabular}

\end{center}
\caption{\label{table-diversity} Effectiveness of RBS and hierarchical copy (HC) mechanism on diversity of comments (\%).} 
\end{table}

\subsection{Emotion Accuracy}
To measure whether the model can generate comments expressing the 
desired emotion, we adopt emotion accuracy
as the agreement between the desired emotion and the predicted emotion by our emotion tagger.
\emph{Acc-C} and \emph{Acc-F} report the coarse-grained and
fine-grained results, respectively.
Of the  four baselines, only Flat-Emo and Hier-Emo are 
emotion-controllable.  The results are shown in Table \ref{table-main}. 

Our first observation is that these two
baselines give similar results on this metric, with Hier-Emo
perform slightly better. Compared to them, 
our model performs significantly better under both the coarse-grained setting and the 
fine-grained setting, with over 3\% absolute improvement. 

More detailed analysis validates the effectiveness of the dynamic fusion 
mechanism and the emotion loss term.  As shown in Table \ref{table-emotion}, using a 
simple fusing method (CCS-Emo) results in drastic drop in emotion accuracy, almost to 
the same level of our baselines. 

\subsection{Diversity}
To measure whether the hierarchical copy mechanism can promote the 
diversity of the generated comments, we report the proportion of novel n-grams in all 
the generated texts, represented as
\emph{Dn}.  This metric has been widely used to evaluate the diversity of generated 
text in a variety of NLG tasks \cite{lin2018global,chen2018fast,li2015diversity}.

As shown in Table \ref{table-main}, all three metrics get improved drastically.
Especially, CCS beats Transformer on D3 by over 7\% (absolute), showing that CCS can 
effectively generate diverse comments compared to strong baselines.

\begin{figure}[]
\centering
\hspace*{-0.5cm}  
\includegraphics[width= 8cm]{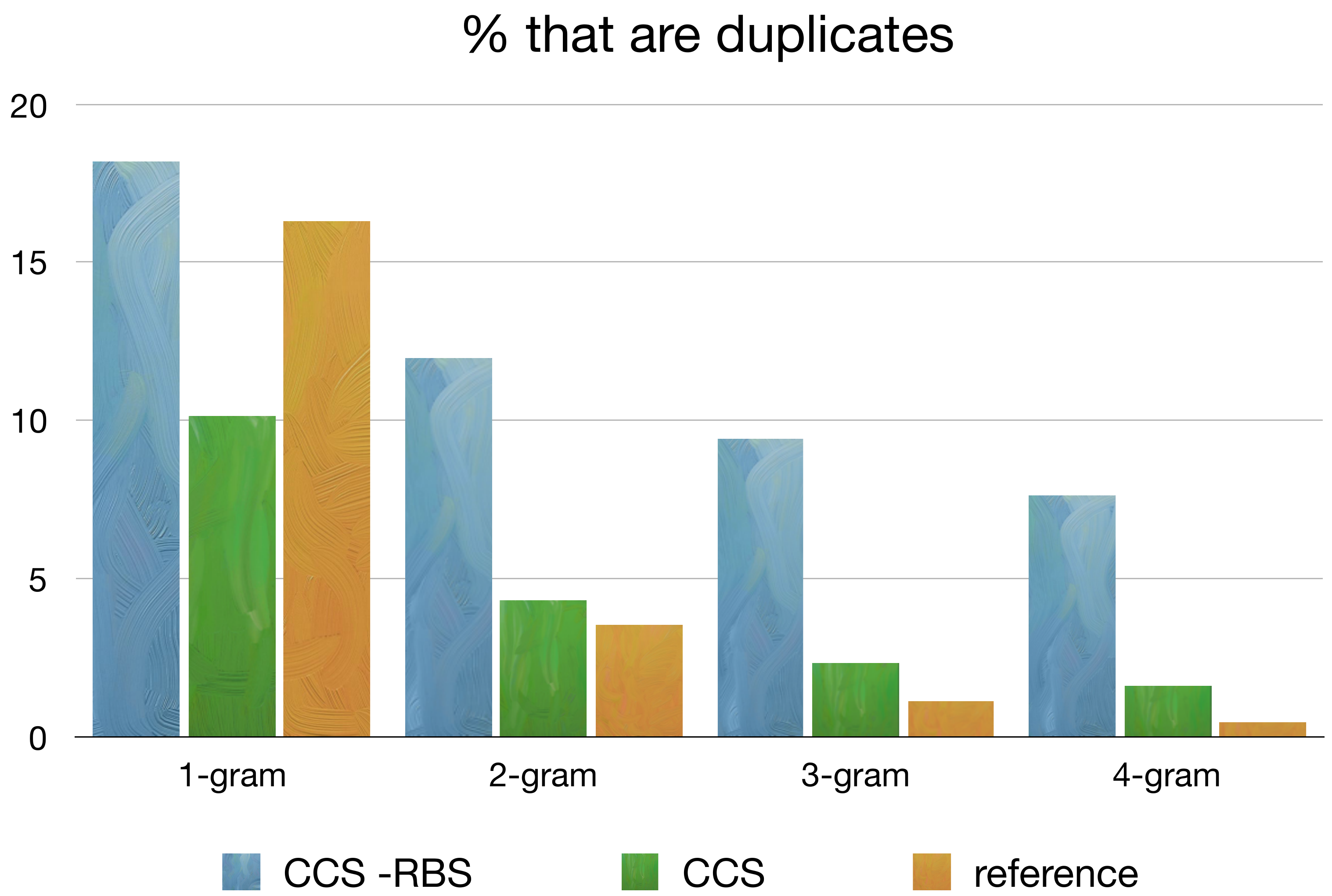}
\caption{Restricted beam search (RBS) helps to mitigate repetition problem. Comments from
normal beam search contain many duplicated n-grams, which our
RBS algorithm produces a similar number as
the reference comments.}
\centering
\label{figure-repetition}
\end{figure}

From Table \ref{table-diversity}, we can see that after removing the hierarchical copy mechanism, the diversity drops significantly (over 2\% absolute for \emph{D2}). Besides, the RBS algorithm also contributes to comments diversity. After removing RBS, the comments diversity drops significantly. To examine the effectiveness of RBS more closely, we report the percentage of repetitive n-grams within comments. From Figure \ref{figure-repetition}, we can observe that the repetition problem (lack of intra-sentence diversity) is rather severe when trained with the normal beam search algorithm (CCS -RBS), with repetitive 3-grams and 4-grams almost ten times more than reference. Despite the 
brevity of our RBS algorithm, the repetition 
problem is almost completely eliminated, with repetitive 3-grams and
4-grams only slightly higher than reference comments.




\begin{figure*}[t]
\centering
\hspace*{-0.5cm}  
\includegraphics[width= 16cm]{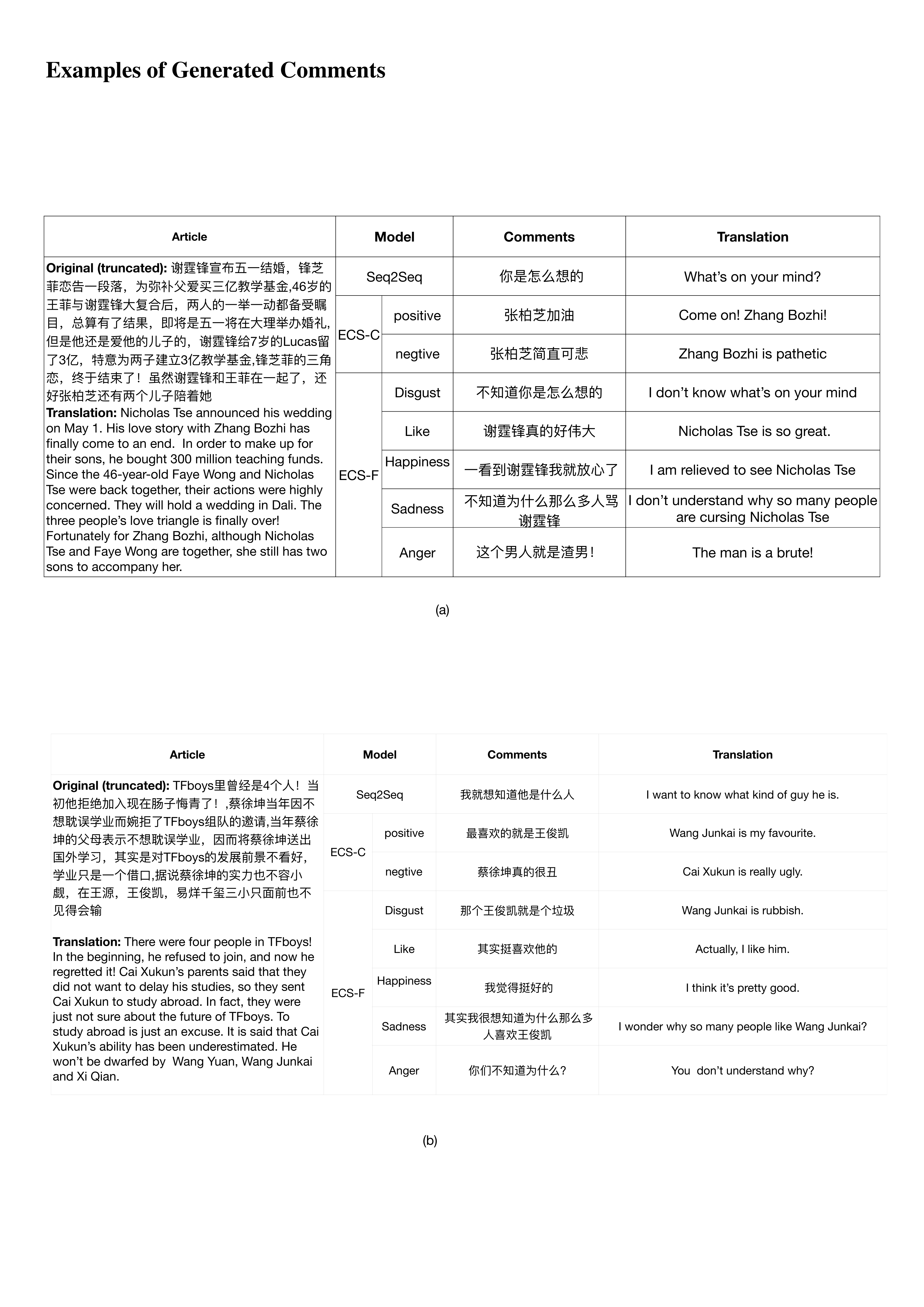}
\caption{Case study.}
\centering
\label{figure-case}
\end{figure*}

\subsection{Case Study}
To gain an insight into how well the emotion is expressed in the generated comments, we provide an example in Figure \ref{figure-case}.
We can see that the Seq2Seq baseline cannot control the expressed emotion, and the comment is not much relevant to the article.
By contrast, our model can generate far more relevant comments.
The main character's name \emph{Bozhi Zhang} and \emph{Nicholas Lse} are directly copied into the comments,
thanks to the hierarchical copy mechanism. 

As for the emotion expressed, we can see that many emotional words
appear in the generated comments, like \emph{pathetic} and \emph{brute}. Besides, there are also other comments that
do not contain any explicit emotional words, yet also express the 
specified emotions. 

However, we do observe some limitations of our model.
The emotion expressed is not always precise. We believe a major reason is that the dataset we use to train our 
emotion tagger is noisy. Besides, some comments are not very informative. We believe a possible solution might be introducing external knowledge. We will leave these to 
future work.




\section{Related Work}
\textbf{Automatic Article Commenting} -
\cite{qin2018automatic} formally proposed the automatic article 
commenting task, along with a large-scale annotated 
article commenting dataset, making data-driven seq2seq approaches to solve this problem viable.  Much research has since focused
on this area \cite{lin2018learning,ma2018unsupervised}.


However, none of these models are controllable, which limits their 
practical application. They also suffer from the general limitations 
of Seq2Seq framework, like repetition and lack of diversity.

\textbf{Controllable Generation} -
\cite{Hu2017TowardCG} is one of the earliest work that consider the
problem of controllable generation. However, they aimed to generate 
text conditioned only on the representation vectors, which is 
significantly different from the task of automatic article 
commenting.  
Our work is close to \cite{Huang2018}, which first introduced 
the emotion factor into dialog generation process. However, as we 
have discussed in the Introduction, the problem of article 
commenting is inherently more challenging. Besides,  they 
simply concatenated the emotion
embedding into the decoder at every time step,  which can be seen as 
a variant of our baseline Flat-Emo. As we have shown in this paper, 
this simple approach gives unsatisfactory results.

\textbf{Copy Mechanism} - \cite{see2017get} proposed the 
Pointer-Generator, which copies words from the source text 
for summarization. However, they regard the input article as 
a flat structure, ignoring the hierarchical structure of document altogether. Based on this observation, we propose the
hierarchical copy mechanism to better suit this task.

\textbf{Restricted Beam Search} -  Our RBS algorithm can be seen as
a soft version of \cite{paulus2017deep}. They set $p(y_t) = 0$ during beam search, when outputting $y_t$ would create a trigram 
that already exists. We, on the other hand, lower the probability of $y_t$ by a certain amount rather than set it to 0. We believe that forcing the decoder to 
\emph{never} output the same trigram more than once is overly 
simplistic, for repetitive trigrams do exist in natural language.

\section{Conclusion}
In this paper, we make the first step towards controlled and diverse article 
commenting. We build two emotional datasets to validate our approach, and propose a 
dynamic fusion mechanism to effectively 
control the expressed emotions of comments. Besides, our model can 
also generate more diverse comments thanks to the hierarchical copy 
mechanism and RBS. Experimental results show that our model beats strong baselines.

\bibliography{anthology,aacl-ijcnlp2020}

\begin{thebibliography}{26}
\expandafter\ifx\csname natexlab\endcsname\relax\def\natexlab#1{#1}\fi

\bibitem[{Chen and Bansal(2018)}]{chen2018fast}
Yen-Chun Chen and Mohit Bansal. 2018.
\newblock Fast abstractive summarization with reinforce-selected sentence
  rewriting.
\newblock \emph{arXiv preprint arXiv:1805.11080}.

\bibitem[{Ghosh et~al.(2017)Ghosh, Chollet, Laksana, Morency, and
  Scherer}]{ghosh2017affect}
Sayan Ghosh, Mathieu Chollet, Eugene Laksana, Louis-Philippe Morency, and
  Stefan Scherer. 2017.
\newblock Affect-lm: A neural language model for customizable affective text
  generation.
\newblock \emph{arXiv preprint arXiv:1704.06851}.

\bibitem[{Hinton et~al.(2012)Hinton, Srivastava, Krizhevsky, Sutskever, and
  Salakhutdinov}]{hinton2012improving}
Geoffrey~E Hinton, Nitish Srivastava, Alex Krizhevsky, Ilya Sutskever, and
  Ruslan~R Salakhutdinov. 2012.
\newblock Improving neural networks by preventing co-adaptation of feature
  detectors.
\newblock \emph{arXiv preprint arXiv:1207.0580}.

\bibitem[{Hochreiter and Schmidhuber(1997)}]{hochreiter1997long}
Sepp Hochreiter and J{\"u}rgen Schmidhuber. 1997.
\newblock Long short-term memory.
\newblock \emph{Neural computation}, 9(8):1735--1780.

\bibitem[{Hu et~al.(2017)Hu, Yang, Liang, Salakhutdinov, and
  Xing}]{Hu2017TowardCG}
Zhiting Hu, Zichao Yang, Xiaodan Liang, Ruslan Salakhutdinov, and Eric~P. Xing.
  2017.
\newblock Toward controlled generation of text.
\newblock In \emph{ICML}.

\bibitem[{Huang et~al.(2018)Huang, Zaane, Trabelsi, and Dziri}]{Huang2018}
Chenyang Huang, Osmar~R Zaane, Amine Trabelsi, and Nouha Dziri. 2018.
\newblock \href {https://doi.org/10.18653/v1/N18-2008} {{Automatic Dialogue
  Generation with Expressed Emotions}}.
\newblock \emph{Naacl}, pages 49--54.

\bibitem[{Kingma and Ba(2014)}]{kingma2014adam}
Diederik~P Kingma and Jimmy Ba. 2014.
\newblock Adam: A method for stochastic optimization.
\newblock \emph{arXiv preprint arXiv:1412.6980}.

\bibitem[{Li et~al.(2015)Li, Galley, Brockett, Gao, and
  Dolan}]{li2015diversity}
Jiwei Li, Michel Galley, Chris Brockett, Jianfeng Gao, and Bill Dolan. 2015.
\newblock A diversity-promoting objective function for neural conversation
  models.
\newblock \emph{arXiv preprint arXiv:1510.03055}.

\bibitem[{Li et~al.(2019)Li, Xu, He, Yan, Wu et~al.}]{li2019coherent}
Wei Li, Jingjing Xu, Yancheng He, Shengli Yan, Yunfang Wu, et~al. 2019.
\newblock Coherent comment generation for chinese articles with a
  graph-to-sequence model.
\newblock \emph{arXiv preprint arXiv:1906.01231}.

\bibitem[{Lin et~al.(2018{\natexlab{a}})Lin, Sun, Ma, and Su}]{lin2018global}
Junyang Lin, Xu~Sun, Shuming Ma, and Qi~Su. 2018{\natexlab{a}}.
\newblock Global encoding for abstractive summarization.
\newblock \emph{arXiv preprint arXiv:1805.03989}.

\bibitem[{Lin et~al.(2018{\natexlab{b}})Lin, Winata, and
  Fung}]{lin2018learning}
Zhaojiang Lin, Genta~Indra Winata, and Pascale Fung. 2018{\natexlab{b}}.
\newblock Learning comment generation by leveraging user-generated data.
\newblock \emph{arXiv preprint arXiv:1810.12264}.

\bibitem[{Luong et~al.(2015)Luong, Pham, and Manning}]{luong2015effective}
Minh-Thang Luong, Hieu Pham, and Christopher~D Manning. 2015.
\newblock Effective approaches to attention-based neural machine translation.
\newblock \emph{arXiv preprint arXiv:1508.04025}.

\bibitem[{Ma et~al.(2018{\natexlab{a}})Ma, Cui, Wei, and
  Sun}]{Ma2018UnsupervisedMC}
Shuming Ma, Lei Cui, Furu Wei, and Xu~Sun. 2018{\natexlab{a}}.
\newblock Unsupervised machine commenting with neural variational topic model.
\newblock \emph{CoRR}, abs/1809.04960.

\bibitem[{Ma et~al.(2018{\natexlab{b}})Ma, Cui, Wei, and
  Sun}]{ma2018unsupervised}
Shuming Ma, Lei Cui, Furu Wei, and Xu~Sun. 2018{\natexlab{b}}.
\newblock Unsupervised machine commenting with neural variational topic model.
\newblock \emph{arXiv preprint arXiv:1809.04960}.

\bibitem[{Meng et~al.(2019)Meng, Li, Sun, Han, Yuan, and Li}]{Meng2019IsWS}
Yuxian Meng, Xiaoya Li, Xiaofei Sun, Qinghong Han, Arianna Yuan, and Jiwei Li.
  2019.
\newblock Is word segmentation necessary for deep learning of chinese
  representations?

\bibitem[{Mikolov et~al.(2013)Mikolov, Chen, Corrado, and
  Dean}]{mikolov2013efficient}
Tomas Mikolov, Kai Chen, Greg Corrado, and Jeffrey Dean. 2013.
\newblock Efficient estimation of word representations in vector space.
\newblock \emph{arXiv preprint arXiv:1301.3781}.

\bibitem[{Paulus et~al.(2017)Paulus, Xiong, and Socher}]{paulus2017deep}
Romain Paulus, Caiming Xiong, and Richard Socher. 2017.
\newblock A deep reinforced model for abstractive summarization.
\newblock \emph{arXiv preprint arXiv:1705.04304}.

\bibitem[{Qin et~al.(2018)Qin, Liu, Bi, Wang, Liu, Hu, Zhao, and
  Shi}]{qin2018automatic}
Lianhui Qin, Lemao Liu, Wei Bi, Yan Wang, Xiaojiang Liu, Zhiting Hu, Hai Zhao,
  and Shuming Shi. 2018.
\newblock Automatic article commenting: the task and dataset.
\newblock \emph{arXiv preprint arXiv:1805.03668}.

\bibitem[{Rose(2016)}]{Rose2016}
Jean Rose. 2016.
\newblock \href {https://doi.org/10.7748/ns.25.22.61.s54} {{Get to the point}}.
\newblock \emph{Nursing Standard}, 25(22):61--61.

\bibitem[{See et~al.(2017)See, Liu, and Manning}]{see2017get}
Abigail See, Peter~J Liu, and Christopher~D Manning. 2017.
\newblock Get to the point: Summarization with pointer-generator networks.
\newblock \emph{arXiv preprint arXiv:1704.04368}.

\bibitem[{Shao et~al.(2017)Shao, Gouws, Britz, Goldie, Strope, and
  Kurzweil}]{shao2017generating}
Louis Shao, Stephan Gouws, Denny Britz, Anna Goldie, Brian Strope, and Ray
  Kurzweil. 2017.
\newblock Generating high-quality and informative conversation responses with
  sequence-to-sequence models.
\newblock \emph{arXiv preprint arXiv:1701.03185}.

\bibitem[{Song et~al.(2019)Song, Zheng, Liu, Xu, and
  Huang}]{song2019generating}
Zhenqiao Song, Xiaoqing Zheng, Lu~Liu, Mu~Xu, and Xuan-Jing Huang. 2019.
\newblock Generating responses with a specific emotion in dialog.
\newblock In \emph{Proceedings of the 57th Conference of the Association for
  Computational Linguistics}, pages 3685--3695.

\bibitem[{Sutskever et~al.(2014)Sutskever, Vinyals, and Le}]{Sutskever2014}
Ilya Sutskever, Oriol Vinyals, and Quoc~V Le. 2014.
\newblock \href {https://doi.org/10.1007/s10107-014-0839-0} {{Sequence to
  sequence learning with neural networks}}.
\newblock \emph{Advances in Neural Information Processing Systems (NIPS)},
  pages 3104--3112.

\bibitem[{Vaswani et~al.(2017)Vaswani, Shazeer, Parmar, Uszkoreit, Jones,
  Gomez, Kaiser, and Polosukhin}]{vaswani2017attention}
Ashish Vaswani, Noam Shazeer, Niki Parmar, Jakob Uszkoreit, Llion Jones,
  Aidan~N Gomez, {\L}ukasz Kaiser, and Illia Polosukhin. 2017.
\newblock Attention is all you need.
\newblock In \emph{Advances in Neural Information Processing Systems}, pages
  5998--6008.

\bibitem[{Wei et~al.(2019)Wei, Lu, Mou, Zhou, Poupart, Li, and
  Jin}]{wei2019neural}
Bolin Wei, Shuai Lu, Lili Mou, Hao Zhou, Pascal Poupart, Ge~Li, and Zhi Jin.
  2019.
\newblock Why do neural dialog systems generate short and meaningless replies?
  a comparison between dialog and translation.
\newblock In \emph{ICASSP 2019-2019 IEEE International Conference on Acoustics,
  Speech and Signal Processing (ICASSP)}, pages 7290--7294. IEEE.

\bibitem[{Zheng et~al.(2018)Zheng, Wang, Chen, and
  Sangaiah}]{zheng2018automatic}
Hai-Tao Zheng, Wei Wang, Wang Chen, and Arun~Kumar Sangaiah. 2018.
\newblock Automatic generation of news comments based on gated attention neural
  networks.
\newblock \emph{IEEE Access}, 6:702--710.

\end{thebibliography}
\bibliographystyle{acl_natbib}

\end{document}